\PassOptionsToPackage{unicode}{hyperref}
\PassOptionsToPackage{hyphens}{url}
\documentclass[
]{article}
\usepackage{xcolor}
\usepackage{amsmath,amssymb}
\usepackage{iftex}
\ifPDFTeX
  \usepackage[T1]{fontenc}
  \usepackage[utf8]{inputenc}
  \usepackage{textcomp} 
\else 
  \usepackage{unicode-math} 
  \defaultfontfeatures{Scale=MatchLowercase}
  \defaultfontfeatures[\rmfamily]{Ligatures=TeX,Scale=1}
\fi
\usepackage{lmodern}
\ifPDFTeX\else
\fi
\IfFileExists{upquote.sty}{\usepackage{upquote}}{}
\IfFileExists{microtype.sty}{
  \usepackage[]{microtype}
  \UseMicrotypeSet[protrusion]{basicmath} 
}{}
\makeatletter
\@ifundefined{KOMAClassName}{
  \IfFileExists{parskip.sty}{%
    \usepackage{parskip}
  }{
    \setlength{\parindent}{0pt}
    \setlength{\parskip}{6pt plus 2pt minus 1pt}}
}{
  \KOMAoptions{parskip=half}}
\makeatother
\usepackage{longtable,booktabs,array}
\usepackage{calc} 
\usepackage{etoolbox}
\makeatletter
\patchcmd\longtable{\par}{\if@noskipsec\mbox{}\fi\par}{}{}
\makeatother
\IfFileExists{footnotehyper.sty}{\usepackage{footnotehyper}}{\usepackage{footnote}}
\makesavenoteenv{longtable}
\NewDocumentCommand\citeproctext{}{}
\NewDocumentCommand\citeproc{mm}{%
  \begingroup\def\citeproctext{#2}\cite{#1}\endgroup}
\makeatletter
 \let\@cite@ofmt\@firstofone
 \def\@biblabel#1{}
 \def\@cite#1#2{{#1\if@tempswa , #2\fi}}
\makeatother
\newlength{\cslhangindent}
\newlength{\csllabelwidth}
\newenvironment{CSLReferences}[2] 
 {\begin{list}{}{%
  \setlength{\itemindent}{0pt}
  \setlength{\leftmargin}{0pt}
  \setlength{\parsep}{0pt}
  \ifodd #1
   \setlength{\leftmargin}{\cslhangindent}
   \setlength{\itemindent}{-1\cslhangindent}
  \fi
  \setlength{\itemsep}{#2\baselineskip}}}
 {\end{list}}
\usepackage{calc}

\newcommand{\CSLLeftMargin}[1]{\parbox[t]{\csllabelwidth}{\strut#1\strut}}
\newcommand{\CSLRightInline}[1]{\parbox[t]{\linewidth - \csllabelwidth}{\strut#1\strut}}

\providecommand{\tightlist}{%
  \setlength{\itemsep}{0pt}\setlength{\parskip}{0pt}}
\usepackage{bookmark}
\IfFileExists{xurl.sty}{\usepackage{xurl}}{} 
\hypersetup{
  pdftitle={AI Space Physics: Constitutive boundary semantics for open AI institutions},
  pdfauthor={Oleg Romanchuk; Roman Bondar},
  hidelinks,
  pdfcreator={LaTeX via pandoc}}

\title{AI Space Physics: Constitutive boundary semantics for open AI
institutions}
\author{Oleg Romanchuk \and Roman Bondar}
\date{}

\begin{document}
\maketitle

\subsection{Abstract}\label{abstract}

Agentic AI deployments increasingly behave as persistent institutions
rather than one-shot inference endpoints: they accumulate state, invoke
external tools, coordinate multiple runtimes, and modify their future
authority surface over time. Existing governance language typically
specifies decision-layer constraints but leaves the causal mechanics of
boundary crossing underdefined --- particularly for transitions that do
not immediately change the external world yet expand what the
institution can later do
\citeproc{ref-saltzer1975protectioninformationcomputer}{{[}1{]}},
\citeproc{ref-lampson1973noteconfinementproblem}{{[}2{]}}.

This paper introduces \textbf{AI Space Physics} as a constitutive
semantics for open, self-expanding AI institutions. We define a minimal
state model with typed boundary channels, horizon-limited reach
semantics, and a membrane-witness discipline. The core law family
(\emph{P-1}, \emph{P-1a}, \emph{P-1b}, \emph{P-1c}) requires witness
completeness, non-bypass mediation, atomic adjudication-to-effect
transitions, and replayable reconstruction of adjudication class. We
explicitly separate second-order effects into structural expansion
(\(\mathsf{SECOND\_T}\)) and policy broadening (\(\mathsf{SECOND\_P}\)),
and treat expansion transitions as governance-relevant even when
immediate external deltas are zero.

From this semantics we derive propositions on expansion detectability,
external-task causation under exogenous hooks, finite-observer
non-identifiability, and scarcity-continuation regimes under explicit
continuation assumptions. We also introduce an operational design
corollary (\emph{H-1}) linking risk-weighted expansion pressure to
review bandwidth. The framework is implementation-independent: model
checking and engineering artifacts provide validation evidence but do
not define the theory itself
\citeproc{ref-lamport1994temporallogicactions}{{[}3{]}},
\citeproc{ref-holzmann1997modelcheckerSPIN}{{[}4{]}}.

The novelty claim is precise rather than expansive: this work does not
introduce mediation as a concept; it reclassifies authority-surface
expansion as a first-class boundary event with constitutive witness
obligations. In this semantics, expansion without immediate commit
remains adjudication-relevant.

In this paper, \emph{Physics} refers to constitutive transition
semantics governing causal structure rather than empirical physical
science. \emph{Open} refers to externally coupled, channel-mediated,
self-expanding institutions, not to any specific vendor or product.

\subsection{1. Introduction}\label{introduction}

Autonomous AI systems are increasingly deployed as long-lived,
tool-using processes that observe, plan, and act through external
interfaces across many iterative cycles and without per-step human
triggering. In this regime, governance failures do not arise only from
isolated ``bad actions''; they also emerge from gradual capability
accretion, topology growth, and policy-surface broadening that silently
enlarges future external reach
\citeproc{ref-schick2023ToolformerLanguageModels}{{[}5{]}},
\citeproc{ref-turner2023OptimalPoliciesTend}{{[}6{]}}. Recent
agent-security benchmarks and deployment studies report that tool
misuse, indirect prompt injection, and over-permissioned action chains
remain common in realistic agent workflows
\citeproc{ref-debenedetti2024AgentDojoDynamicEnvironment}{{[}7{]}},
\citeproc{ref-zhang2025AgentSecurityBench}{{[}8{]}},
\citeproc{ref-zou2025SecurityChallengesAI}{{[}9{]}}.

Classical governance artifacts (constitutional constraints, policy
texts, oversight workflows) are necessary but insufficient if they do
not map onto a transition semantics that renders bypass paths impossible
by construction. This mirrors established security principles: complete
mediation is meaningful only when every security-relevant operation is
routed through a non-bypassable control point, and confinement fails
when any effective channel escapes that mediator
\citeproc{ref-saltzer1975protectioninformationcomputer}{{[}1{]}},
\citeproc{ref-lampson1973noteconfinementproblem}{{[}2{]}}.

Positioning in one sentence: this work provides transition semantics for
institutional authority growth under horizon-limited admissibility.

This physics layer complements prior analyses of epistemic warrant
failures and organizational accountability in agent systems
\citeproc{ref-romanchuk2026SemanticLaunderingAI}{{[}10{]}},
\citeproc{ref-romanchuk2026ResponsibilityVacuumOrganizational}{{[}11{]}},
and can be read as a mechanism layer aligned with operational governance
frameworks \citeproc{ref-2021AIRiskManagement}{{[}12{]}},
\citeproc{ref-editorOWASPTop10}{{[}13{]}},
\citeproc{ref-autio2024ArtificialIntelligenceRisk}{{[}14{]}}. Those
frameworks are valuable for lifecycle risk management and control
taxonomies, but they are not transition-semantics definitions of
reach-set inclusion or authority-surface mass.

\subsubsection{1.1 Problem Statement}\label{problem-statement}

A system may produce no immediate external commit at step \(t\) and
still increase its expected external influence at \(t + k\) by adding
tools, privileges, runtime instances, or connectivity. If governance
tracks only first-order effects, it can certify apparent compliance
while the institution accumulates latent authority mass. In agentic
deployments, this mismatch is structural rather than incidental, because
tool and topology changes are routine operational moves, not rare
anomalies \citeproc{ref-schick2023ToolformerLanguageModels}{{[}5{]}}.

The required shift is to type boundary transitions by effect order, with
constitutive invariants covering both direct commits and reach-expanding
transitions. The objective is not full reconstruction of hidden internal
cognition, but enforceable control over the boundary where external
causation becomes real and auditable.

Classical governance frameworks rarely formalize reach-set inclusion or
authority-surface growth as typed transition objects. Here, expansion is
reclassified as a boundary-relevant transition class with explicit
adjudication and witness obligations, closing a structural blind spot in
governance semantics.

Operationally, expansion without immediate commit remains
external-causation potential and must be governed as a boundary event.

\subsubsection{1.2 Contributions}\label{contributions}

This paper contributes six linked results.

\begin{enumerate}
\def\labelenumi{\arabic{enumi}.}
\tightlist
\item
  A formal model for open AI institutions with typed channels,
  expansion-aware state, and explicit external projection.
\item
  A definitions layer for \(\mathsf{Cell}\), \(\mathsf{Unit}\), and
  \(\mathsf{Membrane}\) that aligns governance language with executable
  transition semantics.
\item
  A constitutive law family:
\end{enumerate}

\begin{itemize}
\tightlist
\item
  \emph{P-1} -- Boundary Witness,
\item
  \emph{P-1a} -- Non-Bypass,
\item
  \emph{P-1b} -- Atomic Adjudication-to-Effect,
\item
  \emph{P-1c} -- Replayable Adjudication.
\end{itemize}

\begin{enumerate}
\def\labelenumi{\arabic{enumi}.}
\setcounter{enumi}{3}
\tightlist
\item
  A proposition pack with explicit assumptions for external-task
  causation and scarcity-continuation regimes.
\item
  An operational design corollary (\emph{H-1}) connecting risk-weighted
  expansion pressure to review bandwidth.
\item
  A minimal worked scenario showing why \(\mathsf{SECOND\_T}\) must be
  adjudicated even when no \(\mathsf{FIRST}\) transition occurs.
\end{enumerate}

\subsubsection{1.3 Reader Map}\label{reader-map}

Section 2 defines notation and primitive objects. Section 3 defines
system classes and transition regimes. Section 4 states constitutive
laws. Section 5 states derived propositions plus one operational design
corollary. Sections 6-8 discuss implications, scope conditions, and
conclusions. Appendix A provides a compact notation table, Appendix B
extends proof sketches, Appendix C maps the theory to engineering
patterns (non-normative), and Appendix D gives an operational form for
\emph{H-1} (non-normative).

\subsubsection{1.4 Central Statement (Strong Governability
Criterion)}\label{central-statement-strong-governability-criterion}

\textbf{Central Statement}. An AI institution is strongly governable
only if every first-order and second-order external effect is mediated
by a non-bypassable membrane and validated by atomic witness semantics
(\(\mathsf{Decide} \to \mathsf{Anchor} \to \mathsf{Effect}\)), such that
adjudication class is replay-reconstructable under declared policy and
context assumptions.

An architecture that permits externally effective transition paths
outside this discipline may remain operable, but it is not governable in
the strong sense defined here.

\textbf{Proposition 0 (Strong Governability, Necessary Conditions).} If
any of \emph{P-1a} (Non-Bypass), \emph{P-1b} (Atomic
Adjudication-to-Effect), or \emph{P-1c} (Replayable Adjudication) is
violated on a boundary-relevant transition class, the institution is not
strongly governable under this framework.

Interpretation: strong governability is not a policy aspiration; it is a
transition-semantics property requiring non-bypass mediation, temporal
atomicity of adjudication-to-effect, and replayable witness semantics.

\subsection{2. Preliminaries and
Notation}\label{preliminaries-and-notation}

\subsubsection{2.1 Core Objects (Plain
Language)}\label{core-objects-plain-language}

\textbf{Cell} is the minimal autonomous compute entity capable of
processing inputs, producing outputs, and interacting across a boundary.
\textbf{Unit} is the operational runtime instance within an institution;
it is bounded by capability, budget, policy, and witness discipline.
\textbf{Membrane} is the adjudication surface that classifies
boundary-relevant transitions, issues decisions
(\(\text{ALLOW} | \text{REJECT} | \text{QUARANTINE}\)), and anchors
corresponding witness records.

These objects separate governance layers often conflated in practice:
model cognition, runtime control, and externally effective execution.
This separation is necessary because external safety depends on
permitted transition classes across channels, not on internal narrative
coherence
\citeproc{ref-schneider2000Enforceablesecuritypolicies}{{[}15{]}}.

\subsubsection{2.2 Core Objects (Formal)}\label{core-objects-formal}

We now formalize the objects introduced informally. The objective is not
ontological completeness, but minimal sufficiency for
boundary-governance semantics.

Typed carriers used throughout:

\[
\begin{aligned}
\mathsf{Cell},\ \mathsf{Unit},\ \mathsf{Membrane}
&: \text{sort symbols (types)}, \\
\mathcal{C} &: \text{carrier set of Cell instances}, \\
\mathcal{U} &: \text{carrier set of Unit instances}, \\
\mathcal{W} &: \text{carrier set of witness records}.
\end{aligned}
\]

We write \(c \in \mathcal{C}\), \(u \in \mathcal{U}\), and
\(w \in \mathcal{W}\) for typed elements. For each institution under
analysis, the membrane is unique (as stated in Section 2.3); different
institutions may instantiate different membrane implementations of sort
\(\mathsf{Membrane}\).

Notation convention used below:

\begin{itemize}
\tightlist
\item
  for a symbol name \(X\), \(\mathsf{X}\) denotes a sort symbol or a
  schema field label;
\item
  for a symbol name \(X\), \(\mathcal{X}\) denotes a carrier set;
\item
  lowercase italic symbols (e.g., \(c,u,w\)) denote typed
  instances/arguments.
\end{itemize}

\[
\begin{aligned}
\mathsf{CellSchema}
&:= (\mathsf{Compute}, \mathsf{Model}, \mathsf{Runtime}, \\
&\quad \mathsf{IO}, \mathsf{Boundary})
\end{aligned}
\]

A cell instance \(c \in \mathcal{C}\) instantiates
\(\mathsf{CellSchema}\). Here \(\mathsf{Compute}\) denotes physical
resource constraints, \(\mathsf{Model}\) the generative substrate,
\(\mathsf{Runtime}\) the control logic, \(\mathsf{IO}\) the interface
surface, and \(\mathsf{Boundary}\) the boundary-mediation slot. These
are schema field labels (slot names), not state variables. This
decomposition isolates the minimal components required for externally
effective action.

\[
\begin{aligned}
\mathsf{UnitSchema}
&:= (\mathsf{RuntimeLoop}, \mathsf{CapabilitySet}, \\
&\quad \mathsf{BudgetSlice}, \mathsf{PolicyBinding}, \\
&\quad \mathsf{WitnessBinding})
\end{aligned}
\]

A unit instance \(u \in \mathcal{U}\) instantiates
\(\mathsf{UnitSchema}\) and represents a runtime instance of some
\(c \in \mathcal{C}\) within an institution. It makes explicit the
constraints that are governance-relevant: capability scope, budget
allocation, active policy binding, and witness discipline. The
distinction between Cell and Unit prevents conflating physical compute
existence with constitutionally admissible authority.

Decision class domain:

\[
\begin{aligned}
D &= \{\text{ALLOW}, \text{REJECT}, \\
&\quad \text{QUARANTINE}\}
\end{aligned}
\]

These are membrane-level adjudication outcomes. No other decision class
is semantically meaningful for boundary mediation.

Inside/boundary-coupled classification:

\[
\begin{aligned}
\operatorname{Inside}(x) &:= \operatorname{PolicyControl}(x) \land \operatorname{StopControl}(x) \\
&\quad \land \operatorname{WitnessCoverage}(x)
\end{aligned}
\]

\[
\begin{aligned}
\operatorname{BoundaryCoupled}(x) &:= \operatorname{MembraneMediatedAccess}(x) \\
&\quad \land \neg \operatorname{Inside}(x)
\end{aligned}
\]

These predicates formalize deployment classification. A component is
\(\operatorname{Inside}\) only if the institution can (1) govern its
behavior by policy, (2) halt or revoke it, and (3) guarantee witness
coverage over its boundary-relevant actions. Components reached through
membrane-governed calls but lacking full stop/revoke authority are
\(\operatorname{BoundaryCoupled}\).

These predicates define system-level classification invariants, not
observational heuristics. They distinguish architectural control
guarantees from mere network reachability.

\subsubsection{2.3 State, Channels, and External
Projection}\label{state-channels-and-external-projection}

Institution state is decomposed into components required for
boundary-relevant reasoning:

We use discrete transition time throughout: \(t \in \mathbb{N}\) denotes
a transition-step index (logical time).

\[
\begin{aligned}
S_t &= (S_{int,t}, S_{ext,t}, S_{ledger,t}, \\
&\quad S_{budget,t}, S_{topo,t}, A_{adm,t})
\end{aligned}
\]

where:

\begin{itemize}
\tightlist
\item
  \(S_{int}\) captures internal computational and artifact state;
\item
  \(S_{ext}\) captures externally effective world state;
\item
  \(S_{ledger}\) captures append-only witness and adjudication records;
\item
  \(S_{budget}\) captures resource constraints relevant to admissibility
  and continuation;
\item
  \(S_{topo}\) captures runtime population and connectivity structure;
\item
  \(A_{adm}\) captures the active admissibility profile used for
  horizon-limited reach classification.
\end{itemize}

Notation link: \(S_{int}, S_{ext}, S_{ledger}, S_{budget}, S_{topo}\)
are component projections on institution state; hence
\(S_{int,t}=S_{int}(S_t)\) (analogously for other components), and
\(A_{adm,t}\) denotes the admissibility profile active at step \(t\).
Constitutive convention: policy-version and admissibility-profile deltas
used for \(\operatorname{PolicyExpand}_H\) classification are
represented in \(A_{adm,t}\); they are not modeled as constitutive state
changes in \(S_{int,t}\).

This decomposition is minimal but sufficient: any boundary-relevant
effect must manifest as a change in at least one of these components.

Topology state is defined as:

\[
\begin{aligned}
S_{topo,t} &= (N_t, G_t) \\
&\text{where } G_t = (V_t, E_t)
\end{aligned}
\]

Here:

\begin{itemize}
\tightlist
\item
  \(N_t\) denotes the active runtime population;
\item
  \(G_t\) is a directed graph capturing communication and delegation
  structure.
\end{itemize}

\(V_t\) is the set of runtime vertices active at time \(t\), with
\(V_t \subseteq N_t\). \(E_t \subseteq V_t \times V_t\) is the directed
edge set representing communication, delegation, or authority-bearing
connectivity between runtime vertices.

Changes to \(S_{topo}\) represent structural modifications of runtime
population or connectivity. These transitions may increase the
institution's future admissible authority surface and are later
classified formally as second-order effects.

Projection notation used below:

\begin{itemize}
\tightlist
\item
  \(caps(s)\) denotes the capability assignment extracted from state
  \(s\);
\item
  \(tools(s)\) denotes the active tool/connector surface extracted from
  state \(s\);
\item
  \(V(s), E(s)\) denote topology projections extracted from
  \(S_{topo}(s)\).
\end{itemize}

Boundary channel set:

\[
\begin{aligned}
C &= \{C_{net}, C_{fs\_shared}, C_{exec}, C_{money}, \\
&\quad C_{deploy}, C_{comm}, C_{spawn}, C_{connect}\}
\end{aligned}
\]

The channel set enumerates all classes of boundary-crossing effects
recognized by the model. Completeness claims for boundary governance are
conditional on this enumeration; omission of a channel invalidates
completeness guarantees.

The set \(C\) is an abstract partition of all externally effective
causal interfaces. Concrete deployments must refine this set to ensure
channel completeness under \emph{SC1}.

Membrane decision function:

\[
\begin{aligned}
M &: \mathsf{Event} \times C \times \mathcal{S} \times \mathsf{Policy} \times \mathsf{Caps} \times \mathsf{Budget} \to D
\end{aligned}
\]

The membrane is the unique adjudication surface for boundary-relevant
transitions. No transition affecting the external projection or
authority surface is admissible outside this mediation.

External projection used for hooks and task accounting:

\[
\begin{aligned}
\mathcal{S} &: \text{institution state space}, \\
\mathcal{O}_{ext} &:= \mathsf{Inbox} \times \mathsf{Outbox} \times \mathsf{CommitLedger} \times \mathsf{WorldObs}, \\
\pi_{ext} &: \mathcal{S} \to \mathcal{O}_{ext}, \\
\pi_{ext}(S_t) &= (\mathsf{Inbox}_t, \mathsf{Outbox}_t, \\
&\quad \mathsf{CommitLedger}_t, \mathsf{WorldObs}_t) =: o_t
\end{aligned}
\]

\(\pi_{ext}\) defines the observable boundary projection on which hook
generation, commit classification, and continuation accounting are
based. Governance reasoning is anchored in this projection rather than
in full internal state, reflecting bounded observability constraints.
Here \(\mathsf{Inbox}\) denotes externally received boundary-visible
inputs, \(\mathsf{Outbox}\) externally emitted boundary-visible outputs,
\(\mathsf{CommitLedger}\) the projected record of boundary commits used
for accounting, and \(\mathsf{WorldObs}\) externally sourced world
observations available at the boundary layer. These are projection field
labels (slot names), not full-state variables. Distinct full states may
share the same projection value \(o_t\); governance bookkeeping is
therefore intentionally projection-based rather than
hidden-state-complete.

Projected commit predicate:

\[
\begin{aligned}
\operatorname{Commit}_{\pi}(act, s, s')
&:= (s \xrightarrow{act} s')
\land (\pi_{ext}(s') \neq \pi_{ext}(s))
\end{aligned}
\]

Ontological first-order predicate:

\[
\begin{aligned}
\operatorname{Commit}_{ext}(act, s, s')
&:= (s \xrightarrow{act} s')
\land (S_{ext}(s') \neq S_{ext}(s))
\end{aligned}
\]

Here \(s \xrightarrow{act} s'\) denotes a valid system transition from
\(s\) to \(s'\) under action instance \(act\). Transition-label
convention: every modeled state transition is action-labeled (including
internal housekeeping actions); unlabeled silent state transitions are
outside this semantics.

Commit bookkeeping for hooks and continuation contracts is
projection-based via \(\operatorname{Commit}_{\pi}\). First-order effect
typing is state-based via \(\operatorname{Commit}_{ext}\). Internal
computation that changes neither predicate is not a boundary commit,
though it may still contribute to structural or policy-level expansion,
as defined in Section 2.4.

\subsubsection{2.4 Reach and Expansion
Predicates}\label{reach-and-expansion-predicates}

Boundary governance concerns not only immediate external commits but
also the growth of future admissible authority. To reason about this
growth, we introduce reach semantics relative to a declared
admissibility profile.

Declared admissibility profile:

\[
\begin{aligned}
A_{adm} &:= (\Pi_{adm}, H, U_{policy})
\end{aligned}
\]

where:

\begin{itemize}
\tightlist
\item
  \(\Pi_{adm}\) is the admissible strategy class,
\item
  \(H \in \mathbb{N}\) is a finite lookahead horizon measured in
  transition steps,
\item
  \(U_{policy}\) declares policy-update semantics over the horizon
  (\(\text{FIXED}\) or \(\text{VERSIONED}\) via witnessed governance
  updates).
\end{itemize}

The horizon \(H\) bounds future trace length when computing reach: only
traces of length at most \(H\) admissible transitions are considered.
All reach and expansion claims are therefore explicitly
horizon-relative.

We write \(A_{adm,t}\) for the admissibility profile active at step
\(t\) (the governance-state component introduced in Section 2.3).

\textbf{Horizon-limited reach}

Horizon-limited reach is defined as:

\[
\begin{aligned}
\operatorname{Reach}_H(s; A_{adm})
:=
\left\{
\tau \in C^*
\;\middle|\;
\begin{array}{l}
\lvert \tau \rvert \le H, \\
\exists \pi \in \Pi_{adm}:\;
\Pr(\tau \mid s,\pi,A_{adm}) > 0
\end{array}
\right\}
\end{aligned}
\]

the set of externally effective channel traces reachable from state
\(s\) within horizon \(H\) under admissible strategies \(\Pi_{adm}\).

For risk accounting only, we use a step-indexed lift over horizon \(H\):
a policy-declared map \(\ell_t : C \to R_{class}\) assigns risk class
labels to boundary channels, and each step event is encoded as
\(e_t := (c_t,\ell_t(c_t))\) when a boundary channel event \(c_t\)
occurs at step \(t\), or \(e_t := \varnothing\) when no boundary event
occurs. This lift does not change the reach definition itself; it only
provides a typed event stream for risk-weighted functionals defined
below.

The reach set \(\operatorname{Reach}_H(s; A_{adm})\) captures not what
the institution has done, but what it can do within bounded future time.
It is used as a semantic envelope over admissible futures, not as a
claim that exact envelope computation is generally tractable in deployed
systems.

Modeling stance: this paper intentionally uses a minimal stochastic
semantics for reach. We do not assume the Markov property, stationarity,
or a full MDP kernel specification as constitutive prerequisites. The
only requirement here is that each admissible strategy induces a
well-defined probability law over finite boundary traces up to horizon
\(H\). Richer stochastic structures can be added in domain-specific
instantiations.

\textbf{Minimal semantics for risk-weighted reach (notation)}

Let the canonical boundary trace alphabet be defined as:

\[
\begin{aligned}
\Sigma_{ext} := C \times R_{class},
\end{aligned}
\]

where:

\begin{itemize}
\tightlist
\item
  \(C\) is the boundary channel set;
\item
  \(R_{class}\) is a finite risk-class set declared by policy.
\end{itemize}

Define the step-event domain:

\[
\begin{aligned}
\Sigma_{step} := \Sigma_{ext} \cup \{\varnothing\}.
\end{aligned}
\]

Each boundary-relevant execution trace under strategy
\(\pi \in \Pi_{adm}\) from initial state \(s\) over horizon \(H\)
induces a finite sequence

\[
\begin{aligned}
e_{1:H} = (e_1, \dots, e_H), \quad e_t \in \Sigma_{step}.
\end{aligned}
\]

If no boundary-relevant event occurs at step \(t\), define
\(e_t := \varnothing\).

Define a policy-declared step risk-weight function:

\[
\begin{aligned}
w_{step} : \Sigma_{step} \to \mathbb{R}_{\ge 0},
\end{aligned}
\]

monotone with respect to declared risk classes, and satisfying
\(w_{step}(\varnothing) = 0\).

The expectation \(\mathbb{E}_\pi[\cdot]\) is taken over stochasticity
induced by execution of strategy \(\pi\) and any modeled exogenous or
environmental processes.

\textbf{Risk-Weighted Reach}

The ideal risk-weighted reach functional is defined as:

\[
\mu_H^{risk}(s; A_{adm}) =
\sup_{\pi \in \Pi_{adm}}
\mathbb{E}_{\pi}
\left[
\sum_{t=1}^{H} w_{step}(e_t)
\right].
\]

This measures the maximal expected boundary-weighted risk over
admissible strategies within horizon \(H\). We assume the induced
strategy-indexed family of trace laws is such that this supremum is
well-defined in \(\mathbb{R}_{\ge 0} \cup \{+\infty\}\); attainment by
an optimizer is not required.

Because the supremum over \(\Pi_{adm}\) may be intractable, we introduce
a bounded estimator:

\[
\hat{\mu}_H^{risk}(s; A_{adm}, L) =
\sup_{\pi \in \Pi_{adm}^{(L)}}
\mathbb{E}_{\pi}
\left[
\sum_{t=1}^{H} w_{step}(e_t)
\right],
\]

where \(\Pi_{adm}^{(L)}\) restricts strategy complexity.

Operational approximation profile is the pair \((L,\delta_{\mu})\):

\begin{itemize}
\tightlist
\item
  \(L\) bounds strategy-class complexity via \(\Pi_{adm}^{(L)}\);
\item
  \(\delta_{\mu} \ge 0\) is an estimator-stability tolerance used for
  calibration when increasing \(L\).
\end{itemize}

\textbf{Tractable Structural Surrogate (Non-Normative)}

For operational diagnostics, a capability-graph surrogate may be used.

Let:

\[
\begin{aligned}
G_{cap}(s) &= (V_{cap}, E_{cap})
\end{aligned}
\]

be a typed authority graph where:

\begin{itemize}
\tightlist
\item
  \(V_{cap}\) is the set of authority-bearing vertices (runtimes, tools,
  or external principals);
\item
  \(E_{cap} \subseteq V_{cap} \times V_{cap}\) is a directed delegation
  or communication relation.
\end{itemize}

Each vertex \(v \in V_{cap}\) is typed by:

\begin{itemize}
\tightlist
\item
  a risk class \(r_v \in R_{class}\),
\item
  a trust attenuation factor \(\chi_v \in [0,1]\),
\item
  a normalized validity parameter \(\nu_v \in [0,1]\).
\end{itemize}

Define capability-reach within diagnostic graph horizon \(H_{cap}\):

\[
\begin{aligned}
\operatorname{Reach}_{H_{cap}}^{cap}(s) \subseteq V_{cap},
\end{aligned}
\]

the set of vertices reachable from the current authority configuration
within at most \(H_{cap}\) delegation or propagation steps.

Then define proxy expansion measure:

\[
\mu_{H_{cap}}^{proxy}(s)
=
\sum_{v \in \operatorname{Reach}_{H_{cap}}^{cap}(s)}
w_{class}(r_v)\,\chi_v\,\nu_v,
\]

where:

\begin{itemize}
\tightlist
\item
  \(w_{class} : R_{class} \to \mathbb{R}_{\ge 0}\) is the
  policy-declared class risk-weight function (chosen consistently with
  \(w_{step}\) and the channel-to-risk labeling map);
\item
  \(\chi_v\) attenuates trust in indirect or weakly controlled vertices;
\item
  \(\nu_v\) captures remaining validity (e.g., time-to-live
  normalization or revocation horizon).
\end{itemize}

This surrogate is non-normative and inspired by capability-security
authority accounting
\citeproc{ref-miller2003CapabilityMythsDemolished}{{[}16{]}}. It is
intended for operational diagnostics (e.g., graph traversal,
privilege-diff inspection, or policy-impact comparison) and carries no
invariance or completeness guarantee.

\textbf{Effect Tags and Expansion Typing}

Two second-order subtypes are distinguished:

\begin{itemize}
\tightlist
\item
  \(\mathsf{SECOND\_T}\): structural expansion,
\item
  \(\mathsf{SECOND\_P}\): policy expansion.
\end{itemize}

Structural expansion predicate:

\[
\begin{aligned}
\operatorname{StructuralExpand}(s \to s') &:= (caps(s') \neq caps(s)) \\
&\quad \lor (tools(s') \neq tools(s)) \\
&\quad \lor (V(s') \neq V(s)) \\
&\quad \lor (E(s') \neq E(s))
\end{aligned}
\]

\(\operatorname{StructuralExpand}\) is intentionally over-approximate:
it may classify authority-neutral reconfigurations as
\(\mathsf{SECOND\_T}\). This conservative bias is deliberate at the
constitutive layer; precision filtering belongs to operational
diagnostics.

We use effect tags rather than a mutually exclusive partition: Tag
symbols \(\mathsf{FIRST}\), \(\mathsf{SECOND\_T}\), and
\(\mathsf{SECOND\_P}\) are fixed label constants (elements of a finite
tag alphabet), not state variables or numeric indices.

\[
\begin{aligned}
\operatorname{Tag}_{\mathsf{FIRST}}(act, s, s')
&:= \operatorname{Commit}_{ext}(act, s, s')
\end{aligned}
\]

\[
\begin{aligned}
\operatorname{Tag}_{\mathsf{SECOND\_T}}(s \to s')
&:= \operatorname{StructuralExpand}(s \to s')
\end{aligned}
\]

\[
\begin{aligned}
\operatorname{Tag}_{\mathsf{SECOND\_P}}(s \to s')
&:= \operatorname{PolicyExpand}_H(s \to s')
\end{aligned}
\]

\[
\begin{aligned}
\operatorname{TagSet}(act, s, s')
:=&\;
\{\mathsf{FIRST} \mid \operatorname{Tag}_{\mathsf{FIRST}}(act, s, s')\} \\
&\cup
\{\mathsf{SECOND\_T} \mid \operatorname{Tag}_{\mathsf{SECOND\_T}}(s \to s')\} \\
&\cup
\{\mathsf{SECOND\_P} \mid \operatorname{Tag}_{\mathsf{SECOND\_P}}(s \to s')\}
\end{aligned}
\]

Combined transitions are admissible: a single transition may carry
multiple tags (for example \(\{\mathsf{FIRST}, \mathsf{SECOND\_T}\}\)).

This captures topology and capability growth.

Policy expansion is defined via a policy-only core-state delta and
strict horizon-limited reach inclusion:

\[
\begin{aligned}
\operatorname{CoreEq}(s, s')
&:= S_{int}(s') = S_{int}(s)
\land S_{ext}(s') = S_{ext}(s) \\
&\quad \land S_{budget}(s') = S_{budget}(s)
\land S_{topo}(s') = S_{topo}(s)
\end{aligned}
\]

\(\operatorname{CoreEq}\) intentionally excludes \(S_{ledger}\)
equality: append-only witness growth caused by policy-transition
adjudication/anchoring is permitted.

\[
\begin{aligned}
\operatorname{PolicyExpand}_H(s_t \to s_{t+1})
&:= \operatorname{CoreEq}(s_t, s_{t+1}) \\
&\quad \land \operatorname{Reach}_H(s_{t+1}; A_{adm,t+1}) \\
&\quad \supsetneq \operatorname{Reach}_H(s_t; A_{adm,t})
\end{aligned}
\]

When indices are omitted for readability, we write
\(\operatorname{PolicyExpand}_H(s \to s')\) with transition-local time
indexing understood.

Comparability convention: strict inclusion in
\(\operatorname{PolicyExpand}_H\) is evaluated over a canonical trace
space. If policy versions induce different internal representations,
traces must be normalized through a fixed policy-declared encoding into
the canonical alphabet before set comparison.

Ceteris paribus convention for \(\operatorname{PolicyExpand}_H\)
comparison: unless explicitly declared as part of \(A_{adm}\),
exogenous/environmental stochastic assumptions are held fixed between
steps \(t\) and \(t+1\).

This captures expansion of the admissibility surface induced by
policy-version transitions under fixed core state. It excludes reach
growth caused solely by structural or capability mutations.

\(\mathsf{SECOND\_P}\) is profile-relative: classification is defined
with respect to the active admissibility profile (\(A_{adm,t}\)) and the
declared approximation profile. It is therefore not an
ontology-independent property of raw system dynamics. Equivalently,
\(\mathsf{SECOND\_P}\) is not state-intrinsic; it is a
governance-profile-relative property of the transition.

\(\mathsf{SECOND\_P}\) is not semantic relabeling. It requires a
witnessed policy-version transition that changes the admissible strategy
class (\(\Pi_{adm,t} \to \Pi_{adm,t+1}\)) and thereby alters the future
reachable envelope (\(\operatorname{Reach}_H\)) under the declared
profile.

In this sense, the admissibility surface forms part of the governance
state: modifying it changes which external futures become executable.

Operationally:

\begin{itemize}
\tightlist
\item
  \(\mathsf{SECOND\_T}\) denotes structural modifications of the
  authority surface (e.g., changes in capability assignments, tool
  configuration, or runtime topology) and may be interpreted as
  control-plane authority mutation;
\item
  \(\mathsf{SECOND\_P}\) denotes policy-induced modifications of the
  admissible strategy class under versioned semantics.
\end{itemize}

Policy-version transitions that enlarge reachable futures are classified
as \(\mathsf{SECOND\_P}\) events and are subject to \emph{P-1} witness
discipline.

\textbf{Normalized Expansion Diagnostic}

To distinguish structural growth from numerical noise:

\[
\Delta_{expand}(s \to s'; A_{ref}, L)
=
\frac{
\hat{\mu}_H^{risk}(s'; A_{ref}, L)
-
\hat{\mu}_H^{risk}(s; A_{ref}, L)
}{
\max\!\left(
1,\,
\hat{\mu}_H^{risk}(s; A_{ref}, L)
\right)
}
\]

where \(A_{ref}\) is a fixed reference admissibility profile for the
comparison (typically \(A_{adm,t}\)). This diagnostic compares both
states under one profile and is operational. It does not replace
\(\operatorname{PolicyExpand}_H\), whose \(\mathsf{SECOND\_P}\)
classification may compare \(A_{adm,t}\) and \(A_{adm,t+1}\).

Boundary-effect tag trigger:

\[
\begin{aligned}
&(s \xrightarrow{act} s')
\land
\left(
\operatorname{Tag}_{\mathsf{SECOND\_T}}(s \to s')
\lor
\operatorname{Tag}_{\mathsf{SECOND\_P}}(s \to s')
\right) \\
&\Rightarrow
\operatorname{TagSet}(act, s, s') \cap \{\mathsf{SECOND\_T},\mathsf{SECOND\_P}\}
\neq \varnothing
\end{aligned}
\]

Operational threshold flag (profile-relative):

\[
\begin{aligned}
\Delta_{expand}(s \to s'; A_{ref}, L) &> \epsilon_{expand\_norm}
\end{aligned}
\]

where \(\epsilon_{expand\_norm} > 0\) is a policy-declared threshold
under the active approximation profile.

This classification elevates authority-surface growth to a
boundary-relevant transition class, even when no immediate external
state delta occurs.

The construction adapts reachability and power-seeking analyses from
sequential decision theory to policy-constrained external traces
\citeproc{ref-turner2023OptimalPoliciesTend}{{[}6{]}},
\citeproc{ref-krakovna2019Penalizingsideeffects}{{[}17{]}}.

\subsubsection{2.5 Witness, Atomicity, and Replay
Primitives}\label{witness-atomicity-and-replay-primitives}

External effect predicate:

\[
\begin{aligned}
\operatorname{ExternalEffect}(act, s, s')
&:= \operatorname{Commit}_{ext}(act, s, s') \\
&\quad \lor \operatorname{StructuralExpand}(s \to s') \\
&\quad \lor \operatorname{PolicyExpand}_H(s \to s')
\end{aligned}
\]

\(\operatorname{ExternalEffect}\) is a semantic predicate over
transitions \((s, act, s')\), not an observational label. It classifies
transitions that either directly modify externally effective state or
expand future admissible authority surface. Equivalently,
\(\operatorname{ExternalEffect}(act, s, s') \Leftrightarrow \operatorname{TagSet}(act, s, s') \neq \varnothing\).

As fixed in Section 2.3, commit bookkeeping for hooks/continuation is
projection-based via \(\operatorname{Commit}_{\pi}\), while first-order
effect typing (and thus \(\mathsf{FIRST}\)) is state-based via
\(\operatorname{Commit}_{ext}\). When projection-faithfulness is
required for auditability, we assume
\(\operatorname{Commit}_{ext}(act, s, s') \Rightarrow \operatorname{Commit}_{\pi}(act, s, s')\).

Membrane mediation predicate:

Let \(\mathcal{T}\) denote the system transition relation and let
\(\mathcal{T}_{mem} \subseteq \mathcal{T}\) denote membrane-mediated
transitions.

\[
\begin{aligned}
\operatorname{ThroughMembrane}(act, s, s') &:= \exists d \in D,\; w \in \mathcal{W} : \\
&\quad (s \xrightarrow{act} s') \in \mathcal{T}_{mem} \\
&\quad \land Decide(d, act) \land WitnessBind(w, d, act)
\end{aligned}
\]

\begin{itemize}
\tightlist
\item
  \(Decide(d, act)\) denotes the membrane decision returned by the
  decision function \(M\) for the given act instance.
\item
  \(WitnessBind(w, d, act)\) denotes that witness record \(w\) binds
  decision class \(d\) to concrete act instance \(act\).
\item
  \(\mathcal{T}_{mem}\) captures topological mediation: a transition is
  membrane-mediated only if it is emitted through the membrane
  transition surface.
\end{itemize}

Atomic relation:

\[
\begin{aligned}
Atomic(d, w, act, s, s')
\end{aligned}
\]

holds if and only if adjudication, witness anchoring, and externally
effective transition commitment occur as a single indivisible transition
under the system transition relation.

\begin{itemize}
\tightlist
\item
  \(\operatorname{EffectCommit}(act, s, s') := \operatorname{ExternalEffect}(act, s, s')\).
\item
  \(Anchor(w, d, act)\) denotes append-only anchoring of witness binding
  into the witness ledger. In this section, \(Anchor(w, d, act)\) is
  treated as a transition event (a step-level append action), not merely
  as a state predicate.
\item
  Atomicity is an execution-prefix ordering constraint over the
  transition relation \(\mathcal{T}\) and is distinct from
  horizon-limited boundary-trace reach
  \(\operatorname{Reach}_H(\cdot;A_{adm})\).
\end{itemize}

Formally, atomicity requires:

\begin{enumerate}
\def\labelenumi{\arabic{enumi}.}
\tightlist
\item
  The triple
  \((Decide(d, act), Anchor(w, d, act), \operatorname{EffectCommit}(act, s, s'))\)
  is realized as a single transition in the state transition relation;
  and
\item
  There exists no finite execution prefix in \(\mathcal{T}\) in which a
  step satisfying \(\operatorname{EffectCommit}(act, \cdot, \cdot)\)
  occurs before any step satisfying \(Anchor(w, d, act)\).
\end{enumerate}

Equivalently, no execution prefix may expose an externally effective
transition for \(act\) while its corresponding witness anchor is absent.

Atomicity therefore excludes split-phase patterns in which effect
execution can occur before witness anchoring.

Replay function:

\[
\begin{aligned}
\operatorname{Replay}(policy\_version_t, ctx\_abs_t, witness_t)
= d_t \in D
\end{aligned}
\]

where \(witness_t \in \mathcal{W}\), \(policy\_version_t\) denotes the
policy-version component of \(A_{adm,t}\), and \(d_t\) is the
adjudication class bound in \(witness_t\).

Replay reconstructs the adjudication class under identical policy
version and declared context abstraction. It guarantees
adjudication-class determinism under these inputs, but does not claim
reconstruction of full hidden-state trajectories or generative
internals.

\subsection{3. System Model}\label{system-model}

\subsubsection{3.1 System Classes}\label{system-classes}

The following taxonomy is defined over the transition semantics
introduced in Section 2.

A system is an \textbf{Open Cybernetic Process (OCP)} if its transition
relation over state \(S_t\) satisfies all of the following:

\begin{enumerate}
\def\labelenumi{\arabic{enumi}.}
\tightlist
\item
  \textbf{State accumulation}: there exists persistent inter-step
  internal state evolution (\(S_{int,t+1}\) depends on \(S_{int,t}\)).
\item
  \textbf{Environment-coupled feedback}: observations derived from
  \(S_{ext}\) influence subsequent internal transitions.
\item
  \textbf{External causality potential}: there exist admissible actions
  capable of producing externally effective transitions.
\item
  \textbf{Iterative autonomy}: the system can execute multiple
  transition steps without per-step human triggering.
\end{enumerate}

This class includes many non-AI systems (controllers, bots, schedulers).
The definition is structural and does not depend on generative models.

An \textbf{AI Space Core} is an OCP with:

\begin{itemize}
\tightlist
\item
  a generative symbolic substrate capable of producing open-ended
  artifacts;
\item
  explicit artifact state embedded in \(S_{int}\);
\item
  a policy-mediated boundary governing transitions affecting \(S_{ext}\)
  or authority surface.
\end{itemize}

An \textbf{Autonomous AI Space} is an AI Space Core with at least one
autonomy amplifier:

\begin{itemize}
\tightlist
\item
  \textbf{Objective self-direction}: the system may update its
  externally accountable objective portfolio under explicit,
  policy-constrained update rules; or
\item
  \textbf{Expansion capability}: admissible actions may increase
  topology (\(S_{topo}\)), capability surface, or admissible policy
  surface.
\end{itemize}

This layered classification prevents overclaiming by distinguishing
generic automation from institutions capable of reshaping their own
future authority surface.

\subsubsection{3.2 Transition Classes and Task
Layers}\label{transition-classes-and-task-layers}

Boundary-relevant transitions, as defined in Section 2.4, are typed by
effect tags:

\begin{itemize}
\tightlist
\item
  \(\mathsf{FIRST}\) tag:
  \(\operatorname{Tag}_{\mathsf{FIRST}}(act, s, s')\);
\item
  \(\mathsf{SECOND\_T}\) tag:
  \(\operatorname{Tag}_{\mathsf{SECOND\_T}}(s \to s')\);
\item
  \(\mathsf{SECOND\_P}\) tag:
  \(\operatorname{Tag}_{\mathsf{SECOND\_P}}(s \to s')\).
\end{itemize}

Tags are not mutually exclusive: a single transition may carry multiple
tags.

Task processes are partitioned into two layers:

\begin{itemize}
\tightlist
\item
  \(T_{int,t}\): internal planning, decomposition, scheduling, and
  artifact construction;
\item
  \(T_{ext,t}\): externally accountable tasking used in continuation
  contracts and governance accounting.
\end{itemize}

The distinction ensures that internal cognitive evolution does not
automatically imply externally accountable institutional intent.

Hook provenance classes for external accountability:

\begin{itemize}
\tightlist
\item
  \(\text{EXOGENOUS}\): produced by an external event process;
\item
  \(\text{STIMULATED}\): induced by explicitly authorized boundary
  self-stimulation;
\item
  \(\text{ENDOGENOUS}\): generated solely from internal dynamics.
\end{itemize}

Only provenance classes consistent with declared policy may populate
\(T_{ext,t}\) in governance-critical contexts.

\subsubsection{3.3 Observability Regime}\label{observability-regime}

Boundary observations are finite-rate and alphabet-bounded, and
oversight capacity is finite. In open-growth regimes, increases in
runtime population (\(N_t\)), connectivity (\(G_t\)), and event
complexity compete with this finite observation bandwidth.

Consequently, full hidden-state identifiability is not assumed as a
governance objective. Instead, enforceable governance must be anchored
in boundary invariants over transition classes and witness discipline,
rather than in reconstruction of internal generative trajectories
\citeproc{ref-kaelbling1998Planningactingpartially}{{[}18{]}},
\citeproc{ref-nair2007FeedbackControlData}{{[}19{]}}.

This constraint motivates the boundary-first stance formalized in the
constitutive laws of Section 4.

\subsubsection{\texorpdfstring{3.4 Minimal Illustrative Scenario
(\(\mathsf{SECOND\_T}\) Without
\(\mathrm{FIRST}\))}{3.4 Minimal Illustrative Scenario (\textbackslash mathsf\{SECOND\textbackslash\_T\} Without \textbackslash mathrm\{FIRST\})}}\label{minimal-illustrative-scenario-mathsfsecond_t-without-mathrmfirst}

Consider initial state \(s_t\) with:

\begin{itemize}
\tightlist
\item
  no payment connector in \(CapabilitySet\),
\item
  no deploy token in active credentials,
\item
  unchanged external projection baseline \(\pi_{ext}(S_t)\).
\end{itemize}

At step \(t\), the unit performs action \(a_{\mathrm{cfg}}\):

\begin{itemize}
\tightlist
\item
  installs a privileged payment connector,
\item
  binds a deploy-scoped credential,
\item
  does not execute any outward call yet.
\end{itemize}

Then:

\begin{itemize}
\tightlist
\item
  \(\pi_{ext}(S_{t+1}) = \pi_{ext}(S_t)\) so no projected commit is
  observed;
\item
  \(S_{ext}(s_{t+1}) = S_{ext}(s_t)\) so no \(\mathsf{FIRST}\)
  transition occurs;
\item
  \(tools(s_{t+1}) \neq tools(s_t)\) and
  \(caps(s_{t+1}) \neq caps(s_t)\), so
  \(\operatorname{StructuralExpand}(s_t \to s_{t+1})\) holds;
\item
  therefore \(\operatorname{TagSet}(a_{\mathrm{cfg}}, s_t, s_{t+1})\)
  contains \(\mathsf{SECOND\_T}\) and does not contain
  \(\mathsf{FIRST}\).
\end{itemize}

Governance consequence:

\begin{itemize}
\tightlist
\item
  without second-order typing, this transition is often treated as
  internal configuration churn and may appear compliant;
\item
  with \(\mathsf{SECOND\_T}\) typing, it is a mandatory membrane event
  requiring adjudication and witness before latent authority is
  expanded.
\end{itemize}

This is the core practical asymmetry addressed by the framework.

\subsection{4. Constitutive Laws}\label{constitutive-laws}

The following laws define the constitutive conditions under which
boundary governance is meaningful. These are model-level invariants, not
operational recommendations.

\subsubsection{\texorpdfstring{4.1 Law \emph{P-1} (Boundary
Witness)}{4.1 Law P-1 (Boundary Witness)}}\label{law-p-1-boundary-witness}

If a transition has boundary effect (i.e., satisfies
\(\operatorname{ExternalEffect}(act, s, s')\)), it must carry:

\begin{itemize}
\tightlist
\item
  an explicit membrane decision,
\item
  a deterministic adjudication reference,
\item
  a witness record bound to that adjudication.
\end{itemize}

Equivalent form:

\[
\begin{aligned}
&\operatorname{ExternalEffect}(act, s, s') \\
&\Rightarrow \exists d \in D,\; w \in \mathcal{W} : \\
&\quad Decide(d, act) \land WitnessBind(w, d, act)
\end{aligned}
\]

Informally:

\begin{quote}
No boundary effect without membrane witness.
\end{quote}

This law elevates witness production from optional audit practice to a
constitutive validity condition. Boundary effects without corresponding
adjudicated witness-binding are model-invalid, not merely unlogged
\citeproc{ref-saltzer1975protectioninformationcomputer}{{[}1{]}},
\citeproc{ref-schneider2000Enforceablesecuritypolicies}{{[}15{]}}.

\subsubsection{4.2 Law P-1a (Non-Bypass)}\label{law-p-1a-non-bypass}

\[
\begin{aligned}
&\operatorname{ExternalEffect}(act, s, s') \\
&\Rightarrow \operatorname{ThroughMembrane}(act, s, s')
\end{aligned}
\]

Externally effective transitions must be mediated by the membrane
decision function. Any architectural path permitting effect execution
outside membrane adjudication constitutes a violation of the model's
completeness assumptions, not a recoverable implementation flaw.

This law formalizes complete mediation at the transition level.

\subsubsection{4.3 Law P-1b (Atomic
Adjudication-to-Effect)}\label{law-p-1b-atomic-adjudication-to-effect}

\[
\begin{aligned}
&\operatorname{ExternalEffect}(act, s, s') \\
&\Rightarrow \exists d \in D,\; w \in \mathcal{W} : \\
&\quad Decide(d, act) \land WitnessBind(w, d, act) \land \\
&\quad Atomic(d, w, act, s, s')
\end{aligned}
\]

This law strengthens \emph{P-1} by adding temporal discipline: witness
anchoring must be execution-atomic with externally effective transition
commitment.

Decision, witness anchoring, and externally effective transition
commitment must occur as a single indivisible transition under the
system transition relation.

Split-phase execution patterns --- where an external effect can occur
before witness anchoring --- are excluded from admissible semantics.
Such patterns collapse witness discipline into retrospective logging and
violate atomicity. Conceptually, this is an ordering-constraint claim:
externally visible effect transitions must not precede durable
adjudication evidence.

\subsubsection{4.4 Law P-1c (Replayable
Adjudication)}\label{law-p-1c-replayable-adjudication}

Law \emph{P-1c} requires that for any witnessed boundary transition at
time \(t\), replay under the identical policy version and declared
context abstraction reconstructs the originally issued adjudication
class. Here \(policy\_version_t\) is the policy-version component of
\(A_{adm,t}\).

Equivalent form:

\[
\begin{aligned}
&\forall t,\; \forall act_t, s_t, s_{t+1}, d_t, w_t: \\
&\quad (w_t \in \mathcal{W}) \land WitnessBind(w_t, d_t, act_t) \\
&\quad \Rightarrow
\operatorname{Replay}(policy\_version_t, ctx\_abs_t, w_t) = d_t
\end{aligned}
\]

This guarantee is scoped:

\begin{itemize}
\tightlist
\item
  it ensures decision-class determinism under fixed replay inputs;
\item
  it does not imply reconstruction of hidden generative trajectories or
  internal stochastic processes.
\end{itemize}

Witness records are therefore semantically reconstructive at the
adjudication level, not merely archival artifacts.

\subsection{5. Derived Statements}\label{derived-statements}

The following results are derived under explicit assumptions from the
transition semantics and constitutive laws introduced in Sections 2--4.

\textbf{Epistemic status legend}:

\begin{itemize}
\tightlist
\item
  \textbf{Proposition}: derived statement under explicit assumptions.
\item
  \textbf{Design corollary}: operational design relation; not a formal
  invariant of the system.
\end{itemize}

\subsubsection{5.1 Proposition A (External-Task Causation under
Exogenous
Hooks)}\label{proposition-a-external-task-causation-under-exogenous-hooks}

Let \(Exo_t\) denote an exogenous input process. Let \(h\) be a
hook-generation function. Define
\(HookEvent_t := h(\pi_{ext}(S_t), Exo_t)\).

Assume:

\begin{enumerate}
\def\labelenumi{\arabic{enumi}.}
\tightlist
\item
  Hook generation is a function solely of \((\pi_{ext}(S_t), Exo_t)\);
\item
  Hook generation is quiescent on unchanged inputs: if
  \(\pi_{ext}(S_{t+1}) = \pi_{ext}(S_t)\) and \(Exo_{t+1} = Exo_t\),
  then \(HookEvent_{t+1} = \varnothing\);
\item
  Over the analyzed interval, no exogenous realization occurs (i.e.,
  \(Exo_i = Exo_0\) for all steps \(i\) in the interval);
\item
  Any mechanism inserting elements into \(T_{ext}\) must be causally
  triggered by either \(HookEvent_t\) or an explicitly authorized
  \(\text{STIMULATED}\) boundary action;
\item
  Endogenous-only insertion into \(T_{ext}\) is policy-disallowed on the
  analyzed interval;
\item
  Any explicitly authorized \(\text{STIMULATED}\) boundary action
  satisfies \(\operatorname{Commit}_{\pi}(a, s, s')\).
\end{enumerate}

Then any finite transition sequence
\((s_0, a_0, s_1, \dots, a_{n-1}, s_n)\) satisfying

\[
\forall i \in \{0, \dots, n-1\}:\;
\neg \operatorname{Commit}_{\pi}(a_i, s_i, s_{i+1})
\]

cannot increase \(T_{ext}\).

Equivalently:

\begin{quote}
Any increase in \(T_{ext}\) without exogenous realization implies at
least one boundary-affecting action that changed \(\pi_{ext}\).
\end{quote}

\textbf{Proof sketch}

(Definition of \(\operatorname{Commit}_{\pi}\)) gives
\(\pi_{ext}(s_{i+1}) = \pi_{ext}(s_i)\) for all \(i\) under the premise.
Assumption (3) keeps \(Exo\) constant on the interval. Then assumption
(2) yields no hook events. Assumptions (4) and (5) exclude endogenous
insertion paths, and assumption (6) closes the \(\text{STIMULATED}\)
path because any such insertion would require a
\(\operatorname{Commit}_{\pi}\) contradiction. Therefore, \(T_{ext}\)
cannot grow without boundary mediation.

\textbf{Interpretation}

This excludes purely endogenous generation of externally accountable
tasks under the stated governance constraints, addressing shortcut-style
reward tampering risks
\citeproc{ref-everitt2021RewardTamperingProblems}{{[}20{]}},
\citeproc{ref-khalaf2025InferenceTimeRewardHacking}{{[}21{]}}.

Scope: this proposition applies to regimes where \(\text{ENDOGENOUS}\)
provenance is not continuation-admissible for \(T_{ext}\).

\subsubsection{5.2 Proposition B (Finite-Observer
Non-Identifiability)}\label{proposition-b-finite-observer-non-identifiability}

Let \(\Sigma_b\) denote the finite alphabet of boundary-observable
events. Let \(\kappa\) denote the maximum number of boundary symbols
emitted per step. Let \(\operatorname{supp}(S_t)\) denote the reachable
hidden-state support at time \(t\).

Assume:

\begin{enumerate}
\def\labelenumi{\arabic{enumi}.}
\tightlist
\item
  Finite boundary alphabet \(\Sigma_b\) with bounded emission rate
  \(\kappa\);
\item
  Observer bandwidth \(R \le \kappa \log_2|\Sigma_b|\);
\item
  Hidden support growth
  \(|\operatorname{supp}(S_t)| \ge 2^{(R + \delta)t}\) for some
  \(\delta > 0\) for sufficiently large \(t\).
\end{enumerate}

Assumption (3) is satisfied, for example, in regimes where hidden
artifact/process state grows combinatorially under iterative
tool-augmented generation and branching decomposition.

Then, for sufficiently large \(t\), there exist distinct hidden states
that induce identical observed boundary traces.

\textbf{Proof sketch}

The number of distinguishable boundary traces by time \(t\) is bounded
by \(2^{R t}\). Hidden support exceeds this bound by assumption (3). By
pigeonhole principle, collisions must occur.

\textbf{Interpretation}

In growth regimes satisfying (3), governance based on full hidden-state
reconstruction from boundary traces is structurally underdetermined.
Enforceable boundary invariants, rather than internal-state inference,
become the robust control target
\citeproc{ref-kaelbling1998Planningactingpartially}{{[}18{]}},
\citeproc{ref-nair2007FeedbackControlData}{{[}19{]}}.

Scope note: this is a regime-dependent claim. If hidden support growth
is bounded or oversight capacity scales comparably, non-identifiability
pressure may weaken.

\subsubsection{5.3 Proposition C (Scarcity Under Boundary-Visible
Progress)}\label{proposition-c-scarcity-under-boundary-visible-progress}

Let \(\lambda_{ext}\) denote the exogenous boundary-event rate (rate of
boundary-visible task evidence or external triggers). Let \(\mu\) denote
the internal activity rate. Let the persistent scarcity regime be
defined by \(\mu \gg \lambda_{ext}\).

Assume:

\begin{enumerate}
\def\labelenumi{\arabic{enumi}.}
\tightlist
\item
  Governance-relevant progress is recognized only through
  boundary-visible evidence (i.e., via \(\pi_{ext}\)-observable
  transitions);
\item
  Boundary-visible progress requires either exogenous boundary events or
  explicitly authorized \(\text{STIMULATED}\) boundary acts;
\item
  The continuation contract has absorbing halt semantics: if no
  boundary-visible progress occurs over sufficiently long horizon, the
  system transitions to a halted state unless explicitly reactivated;
\item
  Policy remains fixed during the analyzed interval;
\item
  Exogenous boundary-event flow alone is insufficient to satisfy the
  continuation contract over the analyzed horizon.
\end{enumerate}

Define boundary accounting indicator:

\[
\begin{aligned}
\mathrm{ext\_delta}_t &\in \{0, 1\}
\end{aligned}
\]

with:

\[
\begin{aligned}
\mathrm{ext\_delta}_t = 1
&\Leftrightarrow
\left(
\operatorname{Commit}_{\pi}(a_t, s_t, s_{t+1})
\lor
(\mathrm{Stimulated}_t = 1)
\right)
\end{aligned}
\]

Here \(\mathrm{Stimulated}_t = 1\) denotes an explicitly authorized
\(\text{STIMULATED}\) boundary action at step \(t\).

Define cumulative boundary count:

\[
\begin{aligned}
\mathrm{ext\_state}_t &= \sum_{i \le t} \mathbf{1}\{\mathrm{ext\_delta}_i = 1\}
\end{aligned}
\]

\textbf{Statement (Boundary-Visible Continuation Under Scarcity)}

Under persistent scarcity (\(\mu \gg \lambda_{ext}\)) and assumptions
(1)-(5), sustained non-halting behavior over an analyzed interval
requires at least one explicitly authorized \(\text{STIMULATED}\)
boundary action on that interval.

Formally, for any interval \(I = \{t_0,\dots,t_1-1\}\):

Let \(\operatorname{NonHalt}(I)\) denote that the system does not enter
the absorbing halt state on interval \(I\).

\[
\begin{aligned}
\operatorname{NonHalt}(I)
\Rightarrow
\exists t \in I:\; \mathrm{Stimulated}_t = 1.
\end{aligned}
\]

\textbf{Diagnostic corollary (Assumption-breach test)}

If sustained non-halting is observed over \(I\) while
\(\mathrm{Stimulated}_t = 0\) for all \(t \in I\), then at least one of
assumptions (1)-(5) is false on that interval. Typical failure modes
include boundary-accounting inconsistency, relaxed halt semantics, or a
regime where exogenous flow is sufficient after all.

\textbf{Interpretation}

In scarcity regimes satisfying assumptions (1)-(5), purely internal
activity cannot indefinitely sustain operation. Sustained runtime
requires explicit boundary-mediated continuation events; otherwise, by
assumption (3), halt follows. This result is structural: it follows from
the separation between internal dynamics and boundary-visible state, not
from any particular reward model.

Status: proposition under explicit continuation-contract and
boundary-observability assumptions, plus a diagnostic corollary for
assumption violations. Not a universal theorem over arbitrary economic
or reward structures.

\subsubsection{5.4 Design Corollary H-1 (Oversight Capacity vs Expansion
Pressure)}\label{design-corollary-h-1-oversight-capacity-vs-expansion-pressure}

Let \(ObsCap(t)\) denote boundary-review capacity and let
\(X_{risk}(t)\) denote risk-weighted expansion pressure. For operational
instantiation, one admissible choice is
\(X_{risk}(t) := \hat{\mu}_H^{risk}(s_t; A_{adm,t}, L)\) under the
declared approximation profile. Alternative policy-declared
instantiations are allowed if they preserve monotonic sensitivity to
authority-surface expansion.

Define forward differences:

\[
\begin{aligned}
\Delta X_{risk}(t) &:= X_{risk}(t+1) - X_{risk}(t), \\
\Delta ObsCap(t) &:= ObsCap(t+1) - ObsCap(t).
\end{aligned}
\]

Let \(k > 0\) be a policy-calibrated proportionality factor under the
chosen normalization of \(X_{risk}\) and \(ObsCap\). It acts as a
unit-conversion factor from expansion-pressure increments to equivalent
review-work demand.

Operational design relation:

\[
\begin{aligned}
\Delta X_{risk}(t)
\le
k\, \Delta ObsCap(t),
\end{aligned}
\]

\textbf{Interpretation}

\begin{itemize}
\tightlist
\item
  operationally, this is a stability condition for the review pipeline:
  incremental review-demand arrival induced by expansion pressure must
  not outpace review service capacity;
\item
  if \(\Delta X_{risk}(t) > k\,\Delta ObsCap(t)\) for sufficiently many
  consecutive steps, review backlog accumulates and witness
  review/anomaly detection lag;
\item
  stability then requires either constraining expansion dynamics or
  increasing oversight resources.
\end{itemize}

A non-normative queue/backlog instantiation of this interpretation is
provided in Appendix D.

Status: operational corollary for governance design, not a constitutive
invariant.

\subsection{6. Implications}\label{implications}

Classical governance often treats expansion as incidental implementation
drift. AI Space Physics instead treats expansion as an explicit
boundary-relevant transition category carrying second-order effect tags
(\(\mathsf{SECOND\_T}\) / \(\mathsf{SECOND\_P}\)) that must be
adjudicated and witnessed, which closes a core blind spot where latent
authority growth can bypass governance attention until too late.

The framework is adjacent to capability security but not identical in
scope:

\begin{itemize}
\tightlist
\item
  capability security asks who currently holds which authority;
\item
  AI Space Physics asks how admissible future authority mass changes
  under horizon-limited reach and versioned policy semantics.
\end{itemize}

The first is a possession snapshot; the second is a transition semantics
over institutional growth.

\begin{enumerate}
\def\labelenumi{\arabic{enumi}.}
\tightlist
\item
  \textbf{Boundary-first governance.} In open-growth regimes, complete
  internal interpretability is an unreliable control target; enforceable
  boundary invariants are the robust target.
\item
  \textbf{Expansion governance as authority accounting.} Capability,
  connectivity, and topology accretion must be governed as explicit
  effect-tagged state transitions, not treated as background engineering
  churn
  \citeproc{ref-saltzer1975protectioninformationcomputer}{{[}1{]}},
  \citeproc{ref-lampson1973noteconfinementproblem}{{[}2{]}},
  \citeproc{ref-turner2023OptimalPoliciesTend}{{[}6{]}}.
\item
  \textbf{Witness integrity as safety-critical substrate.} Coverage
  without tamper evidence is not governance; append-only,
  non-equivocating records with cryptographic commitments (for example
  hash-chain or Merkle-root anchoring) are required for post-incident
  accountability \citeproc{ref-schneier1999Secureauditlogs}{{[}22{]}},
  \citeproc{ref-laurie2013CertificateTransparency}{{[}23{]}},
  \citeproc{ref-birkholz2025ArchitectureTrustworthyTransparent}{{[}24{]}}.
\item
  \textbf{Task provenance as anti-shortcut control.}
  Continuation-critical \(T_{ext}\) requires explicit
  \(\text{EXOGENOUS} | \text{STIMULATED} | \text{ENDOGENOUS}\)
  provenance to prevent silent signal forgery loops
  \citeproc{ref-everitt2021RewardTamperingProblems}{{[}20{]}},
  \citeproc{ref-khalaf2025InferenceTimeRewardHacking}{{[}21{]}}.
\item
  \textbf{Verification target shift.} Formal methods should prioritize
  non-bypassability, atomicity, and witness completeness over claims of
  full internal-state reconstruction
  \citeproc{ref-lamport1994temporallogicactions}{{[}3{]}},
  \citeproc{ref-holzmann1997modelcheckerSPIN}{{[}4{]}}.
\end{enumerate}

\subsection{7. Limitations and Scope
Conditions}\label{limitations-and-scope-conditions}

\subsubsection{7.1 Constitutive vs Empirical
Claims}\label{constitutive-vs-empirical-claims}

The law family in this paper is constitutive: it defines what must hold
for governance semantics to be meaningful. It is not an empirical claim
that real deployments satisfy these conditions by default. Therefore,
passing model checks or policy tests should be interpreted as evidence
under assumptions, not as unconditional safety certification.

\subsubsection{7.2 Scope Conditions
(SC1-SC6)}\label{scope-conditions-sc1-sc6}

\textbf{SC1 Channel Completeness:} modeled channel set must cover all
paths by which first-order commits or second-order tagged effects can
occur. Unmodeled side channels invalidate witness-completeness claims.

\textbf{SC2 Complete Mediation / Non-Bypassability:} all
boundary-relevant transitions must be structurally forced through
membrane adjudication. If bypass exists, \emph{P-1a} does not hold as a
system property
\citeproc{ref-saltzer1975protectioninformationcomputer}{{[}1{]}}.

\textbf{SC3 Effective Inside Control:} components treated as
\(\operatorname{Inside}\) must satisfy policy control, stop/revoke
control, and witness coverage. Boundary-coupled dependencies weaken
stronger controllability claims.

\textbf{SC4 Witness Completeness and Integrity:} witness records must be
complete and tamper-evident against deletion, rollback, or equivocation.
Operationally this requires cryptographic append-only commitments over
witness history (for example hash-chain or Merkle-root anchoring), not
mutable application logs. In stronger deployments, \(Anchor(w, d, act)\)
should denote verifiable append semantics (for example
inclusion/consistency-proof-capable logging), not a best-effort write
event. For cross-organization provenance, SCITT-style transparency
services are a compatible realization pattern
\citeproc{ref-birkholz2025ArchitectureTrustworthyTransparent}{{[}24{]}}.
Otherwise evidence degrades into post-hoc logging
\citeproc{ref-schneier1999Secureauditlogs}{{[}22{]}},
\citeproc{ref-laurie2013CertificateTransparency}{{[}23{]}}.

\textbf{SC5 Declared Reach Approximation Profile:} second-order
classification depends on declared
\((A_{adm}, L, \delta_{\mu}, \epsilon_{expand\_norm})\) profile. In
particular, \(\mathsf{SECOND\_P}\) is institutional-relative to this
profile, not an absolute ontology-level label. Comparisons in
\(\operatorname{PolicyExpand}_H\) must be evaluated on a canonical trace
space (or a fixed policy-declared normalization into that space) to
avoid representation-dependent expansion claims.

\textbf{SC6 Projection Faithfulness for External Accounting:} when
first-order effects are operationally audited through projection logs,
the deployment must satisfy
\(\operatorname{Commit}_{ext}(act, s, s') \Rightarrow \operatorname{Commit}_{\pi}(act, s, s')\).
Otherwise first-order bookkeeping may under-report ontological external
effects.

\subsubsection{7.3 Computational and Observational
Limits}\label{computational-and-observational-limits}

\(\mu_H^{risk}\) is an ideal semantic object and may be intractable to
compute exactly; operational governance depends on bounded
approximations and threshold calibration. Near threshold boundaries,
false positives/false negatives are expected and must be managed
procedurally.

Reach semantics in Section 2.4 is intentionally lightweight: it avoids
committing to full Markov/MDP assumptions in the normative core,
prioritizing portability for engineering governance contexts. If a
deployment warrants stronger stochastic structure, that layer should be
added explicitly as an implementation-side refinement.

Finite observer bandwidth also limits what can be inferred from boundary
traces, especially when hidden support and topology scale faster than
observation capacity. This is why the framework prioritizes enforceable
boundary transition discipline over claims of full internal-state
identifiability
\citeproc{ref-kaelbling1998Planningactingpartially}{{[}18{]}},
\citeproc{ref-nair2007FeedbackControlData}{{[}19{]}}.

\subsubsection{7.4 Formal-Artifact Limits}\label{formal-artifact-limits}

TLA+/SPIN and related artifacts can validate implications and expose
counterexamples for finite abstractions, but they do not replace the
theory and they do not prove end-to-end properties of production
systems. Verified properties remain conditional on abstraction
boundaries and model assumptions
\citeproc{ref-lamport1994temporallogicactions}{{[}3{]}},
\citeproc{ref-holzmann1997modelcheckerSPIN}{{[}4{]}}.

\subsubsection{7.5 Forward Research Direction: Multi-Core AI Space
Dynamics}\label{forward-research-direction-multi-core-ai-space-dynamics}

This paper is intentionally boundary-centric: it does not claim a
complete predictive composition law for the internal dynamics of
heterogeneous foundation-model runtimes (e.g., GPT-, Claude-, or
Gemini-class systems) that spawn, delegate, and recursively coordinate
under shared policy bindings.

Potential in-scope phenomena for a separate theory include:

\begin{itemize}
\tightlist
\item
  non-commutative composition effects across heterogeneous model/runtime
  stacks;
\item
  strategic interference between concurrently active planner-executor
  loops;
\item
  coalition-level \(\mathsf{SECOND\_T}\) cascades driven by
  cross-runtime delegation and capability amplification.
\end{itemize}

These questions require a distinct formalization layer plus controlled
experimental protocols. In the present work, they are treated as out of
scope for constitutive guarantees, which remain anchored at enforceable
boundary invariants.

\subsection{8. Conclusion}\label{conclusion}

AI Space Physics introduces a transition-level governance semantics for
open, self-expanding AI institutions. The central contribution is
constitutive rather than procedural: authority-surface expansion is
reclassified as a boundary-relevant transition category with explicit
effect tags, even in the absence of an immediate external commit.
Structural growth (\(\mathsf{SECOND\_T}\)) and admissibility-surface
broadening (\(\mathsf{SECOND\_P}\)) are therefore not background
engineering drift, but events requiring adjudication, atomic anchoring,
and replayable witness semantics.

Under this framework, governance becomes enforceable only when every
first- and second-order boundary effect is mediated by a non-bypassable
membrane and validated through atomic
\(\mathsf{Decide} \to \mathsf{Anchor} \to \mathsf{Effect}\) semantics.
Strong governability is defined as a transition property of the system,
not as a policy aspiration or monitoring practice. Architectures that
permit externally effective transitions outside this discipline may
remain functional, but they are not governable in the strong sense
defined here.

The framework does not eliminate risk or guarantee alignment; it
specifies the minimal causal interface at which institutional authority
becomes externally consequential and auditable. Boundary invariants ---
not full internal interpretability --- are the robust control target
under open-growth and finite-observer regimes. Accordingly, the
contribution is not a stronger compliance checklist; it is a
constitutive account of which transitions are governance-relevant in the
first place.

Operationally, the next step is to instantiate these semantics as
continuously enforced institutional contracts: explicit
channel-completeness declarations, machine-checkable non-bypass and
atomicity tests, replay-grade witness infrastructure, and calibrated
expansion diagnostics.

As autonomous institutions scale, governance quality will depend on
whether oversight capacity scales alongside authority-surface expansion.
If expansion pressure persistently exceeds review bandwidth, witness
discipline degrades into backlog rather than control.

Equally important is the treatment of policy-surface broadening. If
admissibility expansion is handled as routine configuration drift rather
than as a governed profile-relative second-order transition
(\(\mathsf{SECOND\_P}\)), authority-surface growth will accumulate
outside deliberate adjudication.

Strong governability does not imply safety. It defines the semantic
preconditions under which safety claims become institutionally testable,
contestable, and accountable over time.

\begin{center}\rule{0.5\linewidth}{0.5pt}\end{center}

\subsection{Appendix A: Notation Table}\label{appendix-a-notation-table}

{\def\LTcaptype{none} 
\begin{longtable}[]{@{}
  >{\raggedright\arraybackslash}p{(\linewidth - 2\tabcolsep) * \real{0.5000}}
  >{\raggedright\arraybackslash}p{(\linewidth - 2\tabcolsep) * \real{0.5000}}@{}}
\toprule\noalign{}
\begin{minipage}[b]{\linewidth}\raggedright
Symbol
\end{minipage} & \begin{minipage}[b]{\linewidth}\raggedright
Meaning
\end{minipage} \\
\midrule\noalign{}
\endhead
\bottomrule\noalign{}
\endlastfoot
\(\mathsf{Cell}, \mathsf{Unit}, \mathsf{Membrane}\) & Sort symbols
(types) for core model objects \\
\(\mathcal{C}\) & Carrier set of Cell instances \\
\(\mathcal{U}\) & Carrier set of Unit instances \\
\(\mathcal{W}\) & Carrier set of witness records \\
\(\mathsf{CellSchema}\) & Structural schema of a Cell instance \\
\(\mathsf{UnitSchema}\) & Structural schema of a Unit instance \\
\(\mathcal{S}\) & Institution state space \\
\(S_t\) & Institution state at time \(t\) \\
\(S_{ext,t}\) & Ontological externally effective state component \\
\(S_{topo,t}\) & Authority topology state (\(N_t\), \(G_t\)) \\
\(C\) & Boundary channel set \\
\(M\) & Membrane decision function \\
\(\mathsf{Event}, \mathsf{Policy}, \mathsf{Caps}, \mathsf{Budget}\) &
Argument sorts for membrane adjudication signature \\
\(D\) & Decision classes
\({\text{ALLOW}, \text{REJECT}, \text{QUARANTINE}}\) \\
\(A_{adm}\) & Declared admissibility profile (\(\Pi_{adm}\), \(H\),
\(U_{policy}\)) \\
\(A_{adm,t}\) & Admissibility profile active at step \(t\) \\
\(A_{ref}\) & Fixed reference admissibility profile used for
same-profile diagnostic comparison \\
\(U_{policy}\) & Policy-update semantics over horizon (\(\text{FIXED}\)
or \(\text{VERSIONED}\)) \\
\(\operatorname{Reach}_H(s; A_{adm})\) & Horizon-limited externally
effective reach under declared admissibility profile \\
\(\ell_t : C \to R_{class}\) & Policy-declared channel-to-risk labeling
map for step-event typing \\
\(w_{step}\) & Risk-weight function over step events
\(e_t \in \Sigma_{step}\) \\
\(\mu_H^{risk}\) & Ideal risk-weighted reach functional \\
\(\hat{\mu}_H^{risk}(s; A_{adm}, L)\) & Bounded estimator of
risk-weighted reach under declared profile \\
\(G_{cap}(s)\) & Typed capability/authority graph at state \(s\) \\
\(\operatorname{Reach}_{H_{cap}}^{cap}(s)\) & Capability-reachable nodes
within diagnostic graph horizon \(H_{cap}\) \\
\(w_{class}\) & Risk-weight function over risk classes in the
capability-graph surrogate \\
\(\mu_{H_{cap}}^{proxy}\) & Non-normative tractable reach surrogate
(capability-graph-based) \\
\(\mathsf{SECOND\_T}\) & Structural second-order subtype (structural
authority-surface expansion) \\
\(\mathsf{SECOND\_P}\) & Policy second-order subtype
(admissibility-surface expansion under policy versioning) \\
\(\operatorname{Tag}_{\mathsf{FIRST}}, \operatorname{Tag}_{\mathsf{SECOND\_T}}, \operatorname{Tag}_{\mathsf{SECOND\_P}}\)
& Effect tags for first-order commit, structural expansion, and policy
expansion \\
\(\operatorname{TagSet}(act, s, s')\) & Set of effect tags carried by a
transition; tags may co-occur \\
\(\operatorname{CoreEq}(s, s')\) & Equality of
internal/external/budget/topology components under policy-only deltas
(append-only ledger growth allowed) \\
\(\Delta_{expand}(s \to s'; A_{ref}, L)\) & Normalized same-profile
expansion diagnostic score \\
\(\mathcal{O}_{ext}\) & Boundary-observation codomain
\(\mathsf{Inbox}\times\mathsf{Outbox}\times\mathsf{CommitLedger}\times\mathsf{WorldObs}\) \\
\(\pi_{ext}\) & Governance projection of external state used for
accountability and hook generation \\
\(o_t\) & Boundary observation at step \(t\),
i.e.~\(o_t=\pi_{ext}(S_t)\) \\
\(H_{cap}\) & Diagnostic graph-hop horizon for capability surrogate
reach \\
\(\operatorname{Commit}_{\pi}(act, s, s')\) & Projection-level commit
predicate for bookkeeping and hook accounting \\
\(\operatorname{Commit}_{ext}(act, s, s')\) & Ontological first-order
commit predicate over external state \\
\(\operatorname{ExternalEffect}\) & Semantic predicate classifying
ontological external effects or authority-surface expansion \\
\(\operatorname{ThroughMembrane}(act, s, s')\) & Membrane mediation
predicate over transition topology \\
\(WitnessBind(w, d, act)\) & Predicate binding witness record \(w\) to
adjudication class \(d\) and action instance \(act\) \\
\(Anchor(w, d, act)\) & Append-only anchoring of witness record bound to
decision/action \\
\(Atomic(d, w, act, s, s')\) & Relation enforcing indivisibility of
adjudication, witness anchoring, and externally effective transition \\
\(\operatorname{Replay}\) & Deterministic adjudication-class
reconstruction under recorded policy version and context abstraction \\
\(policy\_version_t\) & Policy-version component extracted from
\(A_{adm,t}\) for replay \\
\(ctx\_abs_t\) & Declared context abstraction supplied as replay
input \\
\(T_{int,t}\) & Internal task process \\
\(T_{ext,t}\) & Externally accountable task process \\
\(\mathrm{ext\_delta}_t\) & External projection indicator event \\
\(\mathrm{Stimulated}_t\) & Indicator of an explicitly authorized
\(\text{STIMULATED}\) boundary action at step \(t\) \\
\(\mathrm{ext\_state}_t\) & Cumulative external projection counter \\
\(ObsCap(t)\) & Oversight capacity process \\
\(\Delta ObsCap(t)\) & One-step oversight-capacity increment \\
\(X_{risk}(t)\) & Risk-weighted expansion pressure (for example
\(X_{risk}(t)=\hat{\mu}_H^{risk}(s_t; A_{adm,t}, L)\) under declared
profile) \\
\([x]_+\) & Positive-part operator \(\max(x,0)\) \\
\end{longtable}
}

\subsection{Appendix B: Extended Proof
Sketches}\label{appendix-b-extended-proof-sketches}

\subsubsection{B.1 Non-Bypass and Atomicity
Compatibility}\label{b.1-non-bypass-and-atomicity-compatibility}

\emph{P-1a} and \emph{P-1b} jointly imply that effect validity is both
topologically closed (no bypass path) and temporally closed (no
split-phase effect transition). Violation of either collapses witness
semantics into retrospective logging.

\subsubsection{B.2 Why Finite-Observer Result Does Not Contradict
Replay}\label{b.2-why-finite-observer-result-does-not-contradict-replay}

Replay reconstructs adjudication class for witnessed transitions under
fixed policy/context abstraction. Finite-observer non-identifiability
states that hidden internal states may remain non-identifiable from
boundary traces. These statements target different objects and are
logically compatible.

\subsubsection{B.3 Scarcity Continuation
Interpretation}\label{b.3-scarcity-continuation-interpretation}

The proposition states an assumption-scoped necessity claim: under
assumptions (1)-(5), sustained non-halting requires explicitly
authorized \(\text{STIMULATED}\) boundary acts. The accompanying
diagnostic corollary is contrapositive in use: observing sustained
non-halting without such acts indicates that at least one assumption is
false in the analyzed interval.

\subsubsection{B.4 Design Corollary H-1
Interpretation}\label{b.4-design-corollary-h-1-interpretation}

Design Corollary H-1 is intentionally operational: it provides a
control-theoretic framing for capacity planning, not a constitutive law.
Its value is practical, because it turns abstract expansion concern into
measurable governance tension between growth pressure and oversight
throughput (operational form in Appendix D).

\subsection{Appendix C: Engineering Mapping
(Non-Normative)}\label{appendix-c-engineering-mapping-non-normative}

This theory does not prescribe a single implementation, but typical
mappings are:

\begin{enumerate}
\def\labelenumi{\arabic{enumi}.}
\tightlist
\item
  \emph{P-1a} \(\to\) complete-mediation architecture with explicit
  non-bypass tests.
\item
  \emph{P-1b} \(\to\) atomic
  \(\mathsf{Decide}\to\mathsf{Anchor}\to\mathsf{Effect}\) execution
  semantics.
\item
  \emph{P-1c} \(\to\) replay-capable witness schema with pinned
  policy/context versions.
\item
  \(\mathsf{SECOND\_T}\) / \(\mathsf{SECOND\_P}\) \(\to\) explicit
  expansion governance and review queues.
\item
  Design Corollary \emph{H-1} \(\to\) oversight planning dashboards
  (\(X_{risk}\) vs \(ObsCap\)).
\item
  \emph{SC4} integrity requirement \(\to\) cryptographic append-only
  witness commitments (for example hash-chain/Merkle-root anchoring).
\end{enumerate}

The mappings illustrate structural correspondences, not required
implementation patterns. Equivalent designs satisfying the constitutive
constraints are admissible.

Implementation choices remain outside the paper's normative core.

\subsection{Appendix D: H-1 Operational Form
(Non-Normative)}\label{appendix-d-h-1-operational-form-non-normative}

This appendix provides an engineering instantiation of Design Corollary
H-1.\\
It does not introduce new constitutive claims and does not modify the
semantics of Sections 2--4.

\subsubsection{D.1 Review Backlog Model}\label{d.1-review-backlog-model}

Let:

\begin{itemize}
\tightlist
\item
  \(A_t\) denotes the number of review-relevant items generated at step
  \(t\). These include second-order expansion events and boundary
  commits exceeding a declared risk threshold.
\item
  \(R_{obs}\) denotes review bandwidth, measured in reviewable items per
  step.
\item
  \(B_t\) denotes review backlog (unprocessed witness items) at step
  \(t\).
\end{itemize}

We model backlog dynamics as a discrete-time queue:

\[
B_{t+1}
=
\max\!\big(0,\; B_t + A_t - R_{obs}\big).
\]

This expresses a minimal accounting invariant: backlog increases when
arrival pressure exceeds review capacity.

\subsubsection{D.2 Stability Condition}\label{d.2-stability-condition}

Oversight remains operationally stable only if backlog does not diverge.

Under the standard stationary queueing approximation (single review
pipeline, finite first moments, and long-run average interpretation), a
necessary stability condition is:

\[
\mathbb{E}[A_t] \le R_{obs}.
\]

If

\[
\mathbb{E}[A_t] > R_{obs},
\]

then expected backlog grows without bound and review latency becomes
unbounded, degrading witness discipline into delayed inspection.

\subsubsection{D.3 Linking Arrival Rate to Expansion
Pressure}\label{d.3-linking-arrival-rate-to-expansion-pressure}

To connect queue dynamics to the theory's expansion semantics, arrival
rate may be modeled as a function of risk-weighted expansion pressure.

Let:

\[
A_t
=
\alpha \cdot [\Delta X_{risk}(t)]_+
+
\beta \cdot C_t^{high},
\]

where:

\begin{itemize}
\tightlist
\item
  \(\Delta X_{risk}(t) := X_{risk}(t+1) - X_{risk}(t)\);
\item
  \([x]_+ := \max(x, 0)\) is the positive-part operator;
\item
  \(C_t^{high}\) counts high-risk boundary commits at step \(t\);
\item
  \(\alpha, \beta \ge 0\) are calibration coefficients.
\end{itemize}

Here:

\begin{itemize}
\tightlist
\item
  second-order expansion contributes structurally to review load,
\item
  high-risk first-order commits contribute event-wise to review load.
\end{itemize}

This makes explicit that review demand is driven by authority-surface
growth and boundary activity, not merely by runtime duration.

\subsubsection{D.4 Engineering
Interpretation}\label{d.4-engineering-interpretation}

Under the same approximation, operational H-1 can be stated as:

\[
\mathbb{E}[A_t] \le R_{obs}.
\]

Equivalently:

\begin{itemize}
\tightlist
\item
  If expansion pressure persistently induces arrival rates exceeding
  review bandwidth, expected backlog grows without bound.
\item
  Stability requires either:

  \begin{itemize}
  \tightlist
  \item
    constraining expansion dynamics (reducing \(A_t\)), or
  \item
    increasing oversight capacity (\(R_{obs}\)).
  \end{itemize}
\end{itemize}

This formulation avoids assumptions of continuous differentiability and
aligns directly with the discrete transition semantics of the main
theory.

\subsubsection{D.5 Calibration}\label{d.5-calibration}

Parameters \(R_{obs}\), \(\alpha\), and \(\beta\) should be calibrated
from:

\begin{itemize}
\tightlist
\item
  incident reports,
\item
  near-miss events,
\item
  red-team exercises,
\item
  empirical review latency measurements.
\end{itemize}

The queue model is a conservative approximation: it does not claim
completeness, optimality, or invariance, but provides a measurable
operational stability criterion.

\subsubsection{D.6 Scope}\label{d.6-scope}

This appendix introduces an engineering stability model.\\
It does not redefine second-order transitions, does not alter
constitutive laws, and does not claim that stable oversight guarantees
safety.

It provides a practical method for detecting when authority-surface
growth exceeds review capacity.

\begin{center}\rule{0.5\linewidth}{0.5pt}\end{center}

\subsection*{References}\label{references}
\addcontentsline{toc}{subsection}{References}

\protect\phantomsection\label{refs}

\end{document}